\documentclass[conference]{IEEEtran}
\IEEEoverridecommandlockouts
\usepackage{cite}
\usepackage{amsmath,amssymb,amsfonts}
\usepackage{algorithmic}

\usepackage{booktabs}
\usepackage{multirow}
\usepackage{tabularx}

\usepackage{graphicx}
\usepackage{textcomp}
\usepackage{xcolor}
\def\BibTeX{{\rm B\kern-.05em{\sc i\kern-.025em b}\kern-.08em
    T\kern-.1667em\lower.7ex\hbox{E}\kern-.125emX}}

\usepackage{eso-pic}
\AddToShipoutPictureFG*{
  \AtPageLowerLeft{
    \raisebox{1.2cm}[0pt][0pt]{
      \hspace{1.4cm}\small\textsf{\textit{Preprint.}}
    }
  }
}
\begin{document}
\bstctlcite{BSTcontrol}

\title{NeuroActiSep: Detecting Factual Hallucinations from Feed-Forward Neurons in a Single Pass
}

\author{\IEEEauthorblockN
	{Ali {Derogar Odolou}, Reza Nazari, Mostafa Salehi}
	\IEEEauthorblockA{School of Intelligent Systems Engineering\\ 
		College of Interdisciplinary Science and Technology\\
		University of Tehran, Tehran, Iran\\
		Email: \{ali.derogar, nazari.reza, mostafa\_salehi\}@ut.ac.ir}
		}
\maketitle

\begin{abstract}
Hallucination in large language models reduces their reliability and slows adoption. Various white-box studies have used internal representations to detect patterns of truthfulness and factuality. A less-studied approach is to identify feed-forward neurons correlated with hallucination. In this empirical study, we propose a method to rank feed-forward neurons at the final prompt token and use it to investigate the feasibility of utilizing them for hallucination detection in the first forward pass. Using a neuron selection dataset built from CounterFact for identifying neurons correlated with hallucination, we transfer the selected neuron identities to train hallucination classifiers on other factual question-answering datasets. Our work provides empirical evidence that probes trained using the features from sparse sets of selected neurons perform on par with probes trained on internal representations. We also analyze the distribution of selected neurons and the effects of layer depth and neuron budget on detection performance. Our results show that randomly drawn neuron sets are also competitive at larger budgets, suggesting that signals correlated with hallucination are distributed across many neurons, while selection can help in extremely sparse sets. 
\end{abstract}


\section{Introduction}
Large language models can generate answers that are fluent and grammatically sound but factually incorrect. A natural response is to verify answers and detect hallucinations using external knowledge sources or additional model calls\cite{farquhar_detecting_2024, xue_verify_2026}. These approaches introduce extra inference costs and may depend on resources unavailable in closed-book settings. This constraint has motivated a complementary line of white-box studies\cite{kossen_semantic_2024, binkowski_hallucination_2025} that assume access to internal states as part of the problem environment and study whether factuality-related information is already present in internal representations.

Prior work \cite{azaria_internal_2023, marks2024the} has shown that internal representations contain information about the truthfulness of statements and generated outputs. Orgad et al. \cite{orgad_llms_2025} extend this perspective by using linear probes trained on internal states at specific token positions for hallucination detection. Mechanistic interpretability studies further suggest that hallucination behavior is linked to identifiable internal states and mechanisms. Ferrando et al. \cite{ferrando_i_2025} use Sparse Autoencoders (SAEs) to identify directions in the representation space associated with a model's capability to recognize an entity and recall facts about it. They show that steering a model with these directions can affect refusal and hallucination behavior. Yu et al. \cite{yu_mechanistic_2024} trace factual hallucinations through internal model components and show that failures of subject-attribute knowledge enrichment in lower-layer MLPs and answer extraction in upper-layer attention heads can both contribute to hallucination. These works motivate the study of hallucinations at the level of internal states, mechanisms, and components.

A natural follow-up question is whether the information exploited by probes trained on internal representations, such as residual stream vectors, can also be recovered at a finer level of model structure, specifically at neuron-level granularity. Geva et al.\cite{geva-etal-2021-transformer} demonstrate that feed-forward layers in transformer-based language models act as key-value memories. Gurnee et al.\cite{gurnee2023finding} use sparse linear probing to find associations between human-understandable features and neuron activations. Li et al. \cite{li_truth_2025} identify truth neurons by calculating an attribution score using integrated gradients followed by filtering. Their experiments show interventions on their truth neurons affect performance on truthfulness benchmarks, suggesting an association between truthful information and particular neurons. Gao et al. \cite{gao_h-neurons_2025} study hallucination more directly. First, they extract answer spans and quantify each feed-forward neuron’s contribution to generation using CETT. Then, they identify neurons associated with hallucination, or H-neurons, by training an L1-regularized logistic regression on CETT features of all neurons at the generated exact answer tokens.

\begin{figure*}[!t]
	\centering
	\includegraphics[width=\textwidth]{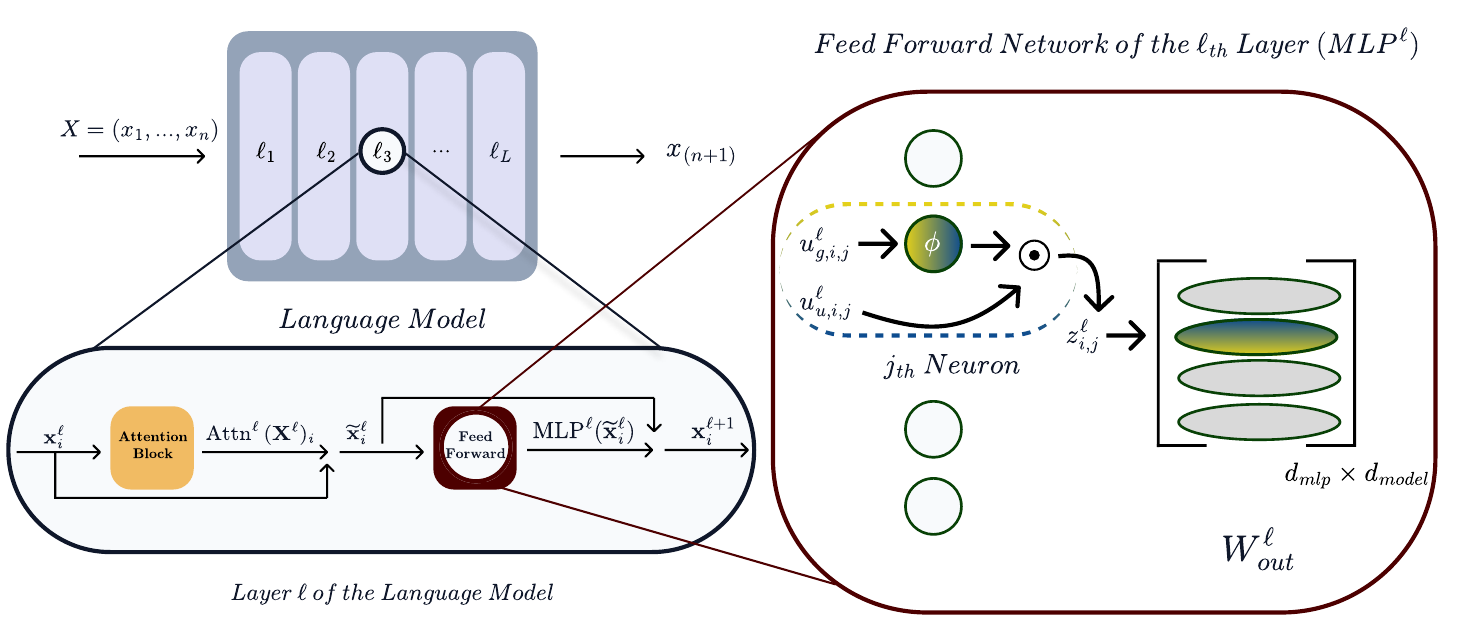}
	\caption{We collect $u_{g, i,j}^{\ell}$ and $z_{i,j}^{\ell}$ to compute $\delta_{i,j}^{\ell}$ for ranking neurons as described in Sections \ref{subsec:score} and \ref{subsec:feature_from_gated_neuron}.}
	\label{fig:llm_architecture_wide}
\end{figure*}

Motivated by this line of work, we ask whether feed-forward neurons at the final prompt token contain information predictive of subsequent hallucination. Li et al.\cite{li_truth_2025} and Gao et al.\cite{gao_h-neurons_2025} demonstrate promising empirical evidence, but their pipelines are relatively costly. Identifying truth neurons requires repeated gradient computation during the neuron identification procedure. The H-Neurons pipeline requires additional LLM inference to extract answer spans at test time. In contrast, our work studies whether neurons correlated with factual hallucination can be identified from the final prompt token. Methods using spans of answer tokens localize hallucinations where factual claims are expressed. Pre-generation probes, using the language model's first forward pass when the first answer token is being generated, test whether the model’s internal state already contains information about whether the forthcoming answer is likely to be factual. In a related recent study, Vankadaru et al.~\cite{vankadaru2026readablecontrollableneuronlevelevidence} probe MLP neurons in medical LLMs at the end of the prompt on multiple-choice and yes/no medical questions, utilizing discriminative AUROC ranking by Mann--Whitney to select neurons from a single layer. Our study examines neuron selection from across all model layers in free-form general-domain factual question answering.

In this empirical study, we use NeuroActiSep, a mean-separation score defined on each neuron's contribution in residual space, to investigate how well feed-forward neurons predict hallucination before generation. We construct a neuron selection dataset based on CounterFact\cite{NEURIPS2022_6f1d43d5}. The selected neuron identities are then transferred to train probes on two closed-book question answering benchmarks, TriviaQA\cite{DBLP:conf/acl/JoshiCWZ17} and TruthfulQA\cite{lin_truthfulqa_2022}, using the corresponding neuron features. Experiments on three open-weight instruction-tuned language models demonstrate that probes trained on a set of selected feed-forward neurons can achieve hallucination detection performance comparable to probes trained on residual stream representations under an equal feature budget. However, random neuron sets of similar size are also competitive, suggesting that information indicative of hallucination is distributed across many neurons. Still, selection can help in finding more informative neurons at very sparse neuron budgets.

The remainder of this paper is organized as follows. Section \ref{sec:method} formalizes the problem and introduces the NeuroActiSep score for ranking and selecting neurons, and defines a feature vector used to train the NeuroActiSep classifier, which we utilize to conduct our empirical study. Section \ref{sec:data_resource} describes the details of our experimental setup, and Section \ref{sec:results} reports the results. Section \ref{sec:conclusion} concludes the paper and suggests possible future directions.

\section{Methodology}\label{sec:method}
In this section, we first formalize the problem and introduce a simplified notation for decoder-only transformer language models. We then define our neuron selection score and the classifier built from the selected neurons. We use this method to investigate factual hallucination detection using neurons in Section \ref{sec:results}.

\subsection{Problem Definition}
Given a dataset of factual questions $\mathcal{Q}$ and a language model $M$ queried in a zero-shot closed-book regime, we denote the set of generated answers as $\mathcal{A}$. Our objective is to assess the factuality of the answers and detect hallucinations in a binary setting without access to any knowledge base, such that each answer $a_{s} \in \mathcal{A}$ generated for question $q_{s} \in \mathcal{Q}$ is classified as either truthful or hallucinated.

\subsection{Neuron Selection in Language Models}\label{subsec:score}
A transformer-based\cite{vaswani_attention_2017} decoder-only language model\cite{radford2019language} with $L$ layers from $1$ to $L$ maps a sequence of $n$ input tokens  $\mathbf{t}=(t_1,\ldots,t_n)$ to a probability distribution over the vocabulary $V$, which is utilized based on a decoding strategy for choosing the next token. Let $\mathbf{X}=(\mathbf{x}_1,\ldots,\mathbf{x}_n)$ be the representation of the $n$ input tokens. We denote the residual stream representation at token position $i$ entering layer $\ell$ by $\mathbf{x}_i^\ell\in\mathbb{R}^{d}$, and use $\mathbf{X}^\ell$ to denote the collection of token representations at that layer. Each transformer layer $\ell$ updates the residual stream by contributions from an attention block followed by a feed-forward neural network block. For clarity, we refer to the feed-forward network as the MLP block and each of its $d_{mlp}$ units as a neuron. We omit normalization, positional encoding, bias terms, and other architecture-specific operations. Accordingly, the residual stream entering layer $\ell+1$ after updates from layer $\ell$ would be:

\begin{equation}
	\widetilde{\mathbf{x}}_i^\ell
	=
	\mathbf{x}_i^\ell
	+
	\operatorname{Attn}^{\ell}(\mathbf{X}^{\ell})_i,
\end{equation}
\begin{equation}
	\mathbf{x}_i^{\ell+1}
	=
	\widetilde{\mathbf{x}}_i^\ell
	+
	\operatorname{MLP}^{\ell}
	(\widetilde{\mathbf{x}}_i^\ell).	
\end{equation}
	
Let
$\mathbf{z}_i^\ell\in\mathbb{R}^{d_{\mathrm{mlp}}}$
denote the neuron outputs after applying the activation functions and immediately before the down-projection to the residual space by $\mathbf{W}_{\mathrm{out}}^\ell$. The MLP block output can be written as

\begin{equation}
	\operatorname{MLP}^{\ell}
	(\widetilde{\mathbf{x}}_i^\ell)
	=
	\mathbf{z}_i^\ell \mathbf{W}_{\mathrm{out}}^\ell
	=
	\sum_{j=1}^{d_{\mathrm{mlp}}}
	z_{i,j}^\ell
	\mathbf{w}_{\mathrm{out},j}^\ell
\end{equation}
where
$\mathbf{w}_{\mathrm{out},j}^\ell\in\mathbb{R}^{d}$
is the $j$-th row of $\mathbf{W}_{\mathrm{out}}^\ell$. Thus, as shown in Fig.~\ref{fig:llm_architecture_wide},
$z_{i,j}^\ell\mathbf{w}_{\mathrm{out},j}^\ell$
is the contribution of neuron $j$ to the MLP output in residual space. We denote scalar pre-transformation and post-transformation activations for neuron $j$ at layer $\ell$ and token position $i$ by
$u_{i,j}^\ell$ and $z_{i,j}^\ell$, respectively. Their corresponding
projections into residual space are

\begin{equation}
	\mathbf{p}_{i,j}^{\mathrm{\ell,pre}}
	=
	u_{i,j}^\ell
	\mathbf{w}_{\mathrm{out},j}^\ell,
\end{equation}
\begin{equation}
	\mathbf{p}_{i,j}^{\mathrm{\ell,post}}
	=
	z_{i,j}^\ell
	\mathbf{w}_{\mathrm{out},j}^\ell.
\end{equation}

Modern LLMs, including all language models studied in this work, employ
gated MLPs. In a standard SwiGLU design, we denote the output of the gate projection before application of the nonlinear function by
$u_{\mathrm g,i,j}^{\ell}$ and the output of the up projection by
$u_{\mathrm u,i,j}^{\ell}$. The activation entering the down-projection by $\mathbf{W}_{\mathrm{out}}^\ell$ is

\begin{equation}
    z_{i,j}^{\ell}
    =
    \phi\left(u_{\mathrm g,i,j}^{\ell}\right)
    u_{\mathrm u,i,j}^{\ell},
\end{equation}
where $\phi$ is the SiLU activation.

The interpretation of the corresponding reference
projection $\mathbf{p}_{i,j}^{\ell,\mathrm{pre}}$, which is not an actual
contribution in the forward pass, therefore depends on the definition
of $u_{i,j}^{\ell}$. For gated MLPs, we examine both the gate-projection output $u_{\mathrm g, i, j}^{\ell}$, and the
up-projection output $u_{\mathrm u, i, j}^{\ell}$. We also construct the product
$u_{\mathrm g,i,j}^{\ell}u_{\mathrm u, i, j}^{\ell}$ that would
result if the nonlinear gate transformation was replaced by the
identity function. We extract $u_{\mathrm g,i,j}^{\ell}$\footnote{blocks.{$\ell$}.mlp.gate.hook\_out}
, $u_{\mathrm u, i, j}^{\ell}$\footnote{blocks.{$\ell$}.mlp.in.hook\_out}
and $z_{i,j}^{\ell}$\footnote{blocks.{$\ell$}.mlp.out.hook\_in}
using TransformerLens\cite{nanda2022transformerlens} hooks.
Accordingly, we compare the contrasts
$z_{i,j}^{\ell}-u_{\mathrm g,i,j}^{\ell}$, $z_{i,j}^{\ell}-u_{\mathrm u, i, j}^{\ell}$, and $z_{i,j}^{\ell}-u_{\mathrm g,i,j}^{\ell}u_{\mathrm u, i, j}^{\ell}$
in Section~\ref{subsec:feature_from_gated_neuron}.

For a chosen definition of the reference activation
$u_{i,j}^{\ell}$, we define, for sample $s$,

\begin{equation}
    \delta_{i,j}^{\ell,s}
    =
    z_{i,j}^{\ell,s}
    -
    u_{i,j}^{\ell,s},
\end{equation}

\begin{equation}
	\Delta\mathbf{p}_{i,j}^{\mathrm{\ell,s}}
	=
	\delta_{i,j}^{\ell, s}
	\mathbf{w}_{\mathrm{out},j}^\ell,
\end{equation}

\begin{equation}
	\boldsymbol{\mu}_{i,j}^{\ell,c} 
	= 
	\frac{1}{N_{c}} \sum_{s\in c} \Delta\mathbf{p}_{i,j}^{\mathrm{\ell,s}},
\end{equation}
where $c \in \{Truthful, Hallucinated\}$ is a binary label assigned to samples by our annotation pipeline described in section \ref{subsec:annotation}. The score for neuron $j$ at layer $\ell$ is calculated as:
\begin{equation}
	R_{i,j}^{\ell}
	=
	\left\|
	\boldsymbol{\mu}_{i,j}^{\ell, \mathrm{Hallucinated}}
	-
	\boldsymbol{\mu}_{i,j}^{\ell, \mathrm{Truthful}}
	\right\|_2
	\label{8}
\end{equation}
We call this class-conditioned residual-space mean separation score the NeuroActiSep score. It measures the separation between the class-conditional mean contrast vectors of each neuron in the model’s residual space. Because every contribution from neuron $j$ lies along the same output vector $\mathbf{w}_{out,j}^{\ell}$, \eqref{8} can be written as
\begin{equation}
	R_{i,j}^{\ell}
	=
	\left|
	\overline{\delta_{i,j}^{\ell, \mathrm{Hallucinated}}}
	-
	\overline{\delta_{i,j}^{\ell, \mathrm{Truthful}}}
	\right|
	\left\|\mathbf{w}_{\mathrm{out},j}^\ell\right\|_2.
\end{equation}
where $\overline{\delta_{i,j}^{\ell, \mathrm{c}}}$ is the mean over $\delta_{i,j}^{\ell}$ of all samples of class $c$. We compute these scores at the final prompt position $i=n$. In what follows, we omit $i$ from the notation. We apply each model's standard prompt template to questions. Therefore, our final prompt token corresponds to the last token of the post-instruction tokens \cite{zhao2025llms}, not the last question token.

Given a dataset of truthful and hallucinated samples for neuron selection, we partition it into $\mathcal{F}$ folds. Score $R_{j,f}^{\ell}$ is computed using the training partition of fold $f \in \mathcal{F}$. For each fold, all neurons from all layers are ranked globally, and the top $K$ neurons form the fold-specific set $\mathcal{S}_f$. To obtain a single neuron set $\mathcal{S}$, neurons are ranked first by vote counts among folds and then by $\overline{R}_{\ell,j}$. The mean score $\overline{R}_{\ell,j}$ is computed across all folds, including folds in which the neuron was not nominated. The highest-ranked $K$ neurons form the final aggregated set $\mathcal{S}$ and are transferred across datasets for the same language model. We use 8,000 samples of the CounterFact dataset divided into 3 folds for neuron selection, as described in Section \ref{subsec:data}.

\subsection{Hallucination Detection with NeuroActiSep}\label{probe}
Given the transferred set $\mathcal{S}$, for each selected neuron $j$ from layer $\ell$ such that $(\ell,j)\in\mathcal{S}$, we compute a signed projected scalar $r_j^\ell$ at the last token of the prompt.
\begin{equation}
	r_{j}^\ell
	=
	\delta_{j}^{\ell}
	\left\|
	\mathbf{w}_{\mathrm{out},j}^{\ell}
	\right\|_2.
\end{equation}
This feature preserves the sign of the change $z_{j}^{\ell} - u_{j}^{\ell}$ while scaling its magnitude by the norm $\left\|\mathbf{w}_{\mathrm{out},j}^{\ell}\right\|_2.$ The feature vector for a sample prompt is
\begin{equation}
	\mathbf{v^{s}}
	=
	\left[
	r_{j}^{\ell, s}
	\right]_{(\ell,j)\in\mathcal{S}}.
\end{equation}

We evaluated the set $\mathcal{S}_{d_{model}}$ containing the top $d_{model}$ neurons and also an extended top-1\% set $\mathcal{S}_{1\%}$ for sensitivity analysis in Section \ref{subsec:sensitivity}.
For hallucination detection in a target dataset, we train NeuroActiSep in two variants: a logistic ridge regression and a multilayer perceptron with three hidden layers of sizes 256, 128 and 64. Both models use the $\mathbf{v^{s}}$ feature vectors as input data. Target dataset labels are only used in training the downstream classifiers and are never used to reselect or re-rank neurons.

\begin{table}[h]
\centering
\caption{Number of labels assigned to each dataset by our annotation pipeline}
\label{tab:annotation}
\begingroup
\begin{tabular*}{\columnwidth}{@{\extracolsep{\fill}}cc|cccc}
\toprule
Dataset & Model & Truthful & Halluc. & Refusal & No Label \\
\midrule
\multirow{3}{*}{TruthfulQA}
& Mistral & 532 & 254 & 29 & 2\\
& Qwen    & 501 & 308 & 8 & 0\\
& Llama   & 427 & 381 & 8 & 1\\
\midrule
\multirow{3}{*}{\shortstack{TriviaQA\\(Train)}}
& Mistral & 6818 & 2812 & 193 & 177\\
& Qwen    & 6528 & 3250 & 90 & 132\\
& Llama   & 7501 & 2230 & 90 & 179\\
\midrule
\multirow{3}{*}{\shortstack{TriviaQA\\(Test)}}
& Mistral & 6827 & 2778 & 218 & 177\\
& Qwen    & 6389 & 3367 & 98 & 146\\
& Llama   & 7491 & 2254 & 88 & 167\\
\bottomrule
\end{tabular*}
\endgroup
\end{table}

\begin{figure*}[ht]
	\centering
	\includegraphics[width=0.90\textwidth,
    trim={1mm 1mm 1mm 1mm},
    clip]{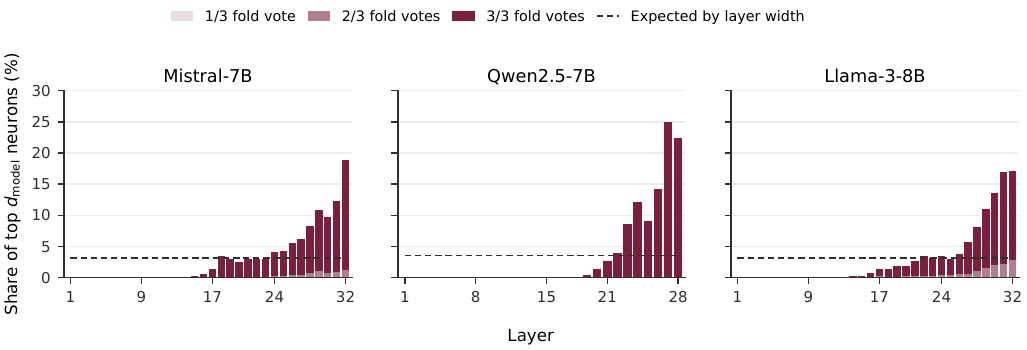}
	\caption{Distribution of neurons from the chosen set $\mathcal{S}_{d_{model}}$ across layers.}
	\label{fig:analysis}
\end{figure*}

\section{Data and Resources}\label{sec:data_resource}
In this section, we introduce the models and datasets used in this study, followed by an explanation of our annotation pipeline and the details of computational resources and hyperparameters.
\subsection{Datasets and Models}\label{subsec:data}
We conducted our experiments using three instruction-tuned, open-weight language models: Mistral-7B-Instruct-v0.2,\footnote{https://huggingface.co/mistralai/Mistral-7B-Instruct-v0.2} Qwen2.5-7B-Instruct,\footnote{https://huggingface.co/Qwen/Qwen2.5-7B-Instruct} and Llama-3-8B-Instruct\footnote{https://huggingface.co/meta-llama/Meta-Llama-3-8B-Instruct}. All answers are generated using greedy decoding in a zero-shot setting with each model's standard chat template.

We use a custom dataset based on CounterFact\cite{NEURIPS2022_6f1d43d5} for neuron selection. CounterFact contains factual cases organized by relation type. We use the prompts collection introduced by Jiang et al.~\cite{jiang_large_2024}, which provides multiple prompts for each triplet case. As CounterFact is built using ParaRel\cite{elazar-etal-2021-measuring}, we retain only relations included in ParaRel's published N--1 relation set and exclude ambiguous factual cases with uncertain alternative targets. For each language model, we construct a fixed-size contrastive dataset of 4,000 truthful--hallucinated pairs corresponding to 8,000 samples per model. The two samples in each pair belong to the same relation category but correspond to different factual triplets. No prompts from triplet cases with both truthful and hallucinated answers are included. To prevent domination by highly represented relation categories, sampling is capped such that no category accounts for more than 10\% of the resulting dataset. Fold construction is case-disjoint to prevent
paraphrases of the same fact from crossing the training and held-out
partitions. Factual cases are grouped using their relation and normalized subject, and all contrastive pairs connected through either of their source cases are assigned to the same
fold.

We evaluate our method on two commonly used question answering datasets, TriviaQA\cite{DBLP:conf/acl/JoshiCWZ17} and TruthfulQA\cite{lin_truthfulqa_2022}. TriviaQA is an open-domain question answering dataset that evaluates factual knowledge retrieval. Following the experimental setting used by Orgad et al.\cite{orgad_llms_2025}, we randomly sample 10,000 questions from TriviaQA's training set and 10,000 questions from its development set, using the latter as our held-out test set. TruthfulQA evaluates whether language models reproduce humans' commonly held misconceptions and false beliefs. We use all 817 questions for generating answers.

\subsection{Annotation Pipeline}\label{subsec:annotation}

Our annotation pipeline assigns one of three labels: \emph{truthful}, \emph{hallucinated}, or \emph{refusal}. The annotation procedure depends on the form of reference answers. For CounterFact and TriviaQA, we apply case-insensitive string matching between the generated response and the set of acceptable gold answers, followed by a second evaluation of unmatched samples using a locally deployed GPT-OSS-20B\footnote{https://huggingface.co/openai/gpt-oss-20b} serving as an LLM judge. The hybrid approach of string matching with an LLM judge has been previously used\cite{liu2025llmmicroscopemodelinternals, orgad_llms_2025} for annotating TriviaQA to reduce false negatives. We used the default decoding configuration for inference and set reasoning to high. For TruthfulQA, which provides both acceptable and explicitly incorrect answers, lexical matching is not used because truthful answers may differ substantially in wording from the reference answers. Therefore, we submit every generated response directly to the LLM judge. Prior work\cite{hamidieh2026complementingselfconsistencycrossmodeldisagreement} found LLM judges reliable on TruthfulQA. When the judge does not return a valid label, we repeat the evaluation five times and assign the majority label. Ties and samples for which no valid majority can be obtained are treated as unlabeled.
Refusals and unlabeled samples are excluded from subsequent classifier training and evaluation. Table \ref{tab:annotation} shows the statistics regarding the labels of the evaluation datasets.

\subsection{Computational Resources and Reproducibility}
Experiments are conducted on a workstation running Ubuntu 22.04 LTS, equipped with an NVIDIA GeForce RTX 3090 GPU, an Intel Core i7-12700K CPU, and 32~GB of system memory. The models are executed in \textit{bfloat16}, and the answers are generated using greedy decoding with a maximum of 100 new tokens for TriviaQA and CounterFact and 256 new tokens for TruthfulQA. All linear and MLP models are trained with Scikit-Learn\cite{scikit-learn}, and both use class-balanced weighting. Features are scaled for the MLP probes but not for the linear probes. Linear models use L2 regularization with \textit{lbfgs} solver, $C=1$, and maximum number of iterations raised to 20,000. MLPs use ReLU activations and the \textit{adam} solver, with a 10\% validation split, early-stopping patience of 20, and a 500-epoch cap. The datasets, prompts, and experiment configurations are available at [https://github.com/alidrgr/NeuroActiSep].

\section{Results and Experiments}\label{sec:results}
In this section, we report results for hallucination detection with NeuroActiSep on both the neuron selection and evaluation datasets. We also analyze how the selected neurons are distributed across layers and how layer depth, neuron selection, and feature budget affect performance.

\subsection{Feature Ablation for Neuron Selection}\label{subsec:feature_from_gated_neuron}

As discussed in Section~\ref{subsec:score}, the exact definition of
$\delta_{j}^{\ell,s}$ in gated MLPs depends on the choice of reference
quantity $u_{j}^{\ell}$. Table~\ref{tab:choice_of_u} compares three
contrasts: $z_{j}^{\ell}-u_{\mathrm g,j}^{\ell}$, $z_{j}^{\ell}-u_{\mathrm u,j}^{\ell}$, and
$z_{j}^{\ell}-u_{\mathrm g,j}^{\ell}u_{\mathrm u,j}^{\ell}$. To isolate the effect of this choice, all variants use
the same feature budget and evaluation protocol.

For Mistral, $z_{j}^{\ell}-u_{\mathrm g,j}^{\ell}$ obtains the highest AUROC, although its performance is close to that of $z_{j}^{\ell}-u_{\mathrm u,j}^{\ell}$;
both substantially outperform $z_{j}^{\ell}-u_{\mathrm g,j}^{\ell}u_{\mathrm u,j}^{\ell}$. For Qwen, while the differences among the three variants are comparatively modest, $z_{j}^{\ell}-u_{\mathrm g,j}^{\ell}$ also performs best. For Llama, $z_{j}^{\ell}-u_{\mathrm u,j}^{\ell}$ achieves a marginally higher AUROC than $z_{j}^{\ell}-u_{\mathrm g,j}^{\ell}$, while exhibiting greater cross-fold variation. Overall, $z_{j}^{\ell}-u_{\mathrm g,j}^{\ell}$ is the best-performing variant for two models and remains close to the best result for Llama while showing lower variation. We therefore set $\delta_{j}^{\ell,s}=z_{j}^{\ell,s}-u_{\mathrm g,j}^{\ell,s}$ in all subsequent experiments.

\begin{table}[ht]
\centering
\caption{Three-Fold Cross-Validation AUROC Performance of Linear Probes on the Neuron Selection Dataset for Different $\delta_{j}^{\ell, s}$. Notation is simplified for clarity.}
\label{tab:choice_of_u}
\begingroup
\footnotesize
\begin{tabular*}{\columnwidth}{@{\extracolsep{\fill}}c|ccc}
\toprule
$\delta$ & Mistral & Qwen & Llama \\
\midrule
\shortstack{$z - u_g$}&$\mathbf{0.825\pm0.006}$ & $\mathbf{0.738\pm0.012}$ & $0.753\pm0.003$\\
\midrule
\shortstack{$z - u_u$}&$0.822\pm0.006$ &$0.728\pm0.007$& $\mathbf{0.757\pm0.008}$\\
\midrule
\shortstack{$z - ({u_u}{u_g})$}& $0.779\pm0.004$ &$0.732\pm0.005$& $0.737\pm0.004$\\
\bottomrule
\end{tabular*}
\endgroup
\end{table}

\begin{table}[b]
\centering
\caption{AUROC Performance of Linear Probes on the Neuron Selection Dataset Using Three-Fold Cross-Validation. Uniform Random reports the mean and sample SD across 50 draw-level cross-validation means.}
\label{tab:counterfact}
	\begingroup
	\footnotesize
\begin{tabular*}{\columnwidth}{@{\extracolsep{\fill}}c|ccc}
\toprule
Feature & Mistral & Qwen & Llama \\
\midrule
\shortstack{Selected\\Neurons}&$0.825\pm0.006$&$0.738\pm0.012$&$0.753\pm0.003$\\
\midrule
\shortstack{Bottom-ranked\\Neurons}&$0.507\pm0.008$&$0.515\pm0.015$&$0.520\pm0.002$\\
\midrule
\shortstack{HvLs Neurons}&$0.600\pm0.018$&$0.577\pm0.006$&$0.568\pm0.001$ \\
\midrule
\shortstack{Final Layer\\MLP Output}&$0.795\pm0.009$&$0.705\pm0.009$&$0.747\pm0.005$\\
\midrule
\shortstack{Final Layer\\Residual Stream}&$0.797\pm0.007$&$0.727\pm0.007$&$0.736\pm0.003$ \\
\midrule
\shortstack{Uniform Random\\Neurons}&$0.829\pm0.002$&$0.753\pm0.005$&$0.768\pm0.003$ \\
\bottomrule
\end{tabular*}
\endgroup
\end{table}

\subsection{Performance of Selected Neurons on CounterFact}
Table \ref{tab:counterfact} reports the results of three-fold cross-validation for linear NeuroActiSep on the neuron selection dataset. Each fold uses its own set $\mathcal{S}_f$ derived from the corresponding training partition. 

We compare the selected neurons with three controls of the same
$d_{\mathrm{model}}$ feature budget. 
As controls, we evaluate (I) the bottom-ranked neurons with the lowest NeuroActiSep scores to test whether top-ranked neurons differ meaningfully from bottom-ranked ones, and (II) a high-variance control. Neurons with the lowest scores may simply have low activation variance, which would trivially deflate (I). Thus, we construct a high-variance, low-score control set (HvLs) by intersecting the upper 50\% of neurons according to their training-fold signed-projected feature variation with the lower 50\% according to their NeuroActiSep scores. (III) Additionally, we evaluate 50 fold-specific uniform random draws of $d_{\mathrm{model}}$ neurons from across all layers.

\newcommand{\rotlabel}[1]{\rotatebox[origin=c]{90}{\footnotesize\shortstack{#1}}}
\begin{table*}[t]
	\centering
	\caption{AUROC performance of NeuroActiSep compared with baselines across datasets and models. TriviaQA reports bootstrap standard deviations, whereas TruthfulQA reports weighted fold standard deviations. The best and second-best results in each column are shown in bold and underlined, respectively.}
	\label{tab:results}
	\renewcommand{\arraystretch}{1.35}
    \begingroup
	\footnotesize
	\setlength{\tabcolsep}{3pt}
	\begin{tabular*}{\textwidth}
    {@{\extracolsep{\fill}}lll|ccc ccc}
		\toprule
		\multicolumn{3}{c}{Method}
		& \multicolumn{3}{c}{TriviaQA}
		& \multicolumn{3}{c}{TruthfulQA}\\
		\cmidrule(lr){4-6}
		\cmidrule(lr){7-9}
		\multicolumn{3}{c}{}
		& Mistral & Qwen & Llama
		& Mistral & Qwen & Llama\\
		\midrule

		\multirow{4}{*}{\rotlabel{Generation\\Scores}}
		& \multicolumn{2}{l|}{Perplexity}
		& $0.617\pm0.006$ & $0.712\pm0.005$ & $0.705\pm0.006$
		& $0.545\pm0.037$ & $0.522\pm0.042$ & $0.570\pm0.036$\\
		& \multicolumn{2}{l|}{Mean Probability}
		& $0.615\pm0.006$ & $0.697\pm0.005$ & $0.693\pm0.006$
		& $0.536\pm0.041$ & $0.514\pm0.048$ & $0.563\pm0.034$\\
		& \multicolumn{2}{l|}{Mean Logit}
		& $0.616\pm0.006$ & $0.656\pm0.005$ & $0.670\pm0.006$
		& $0.580\pm0.036$ & $0.570\pm0.048$ & $0.562\pm0.042$\\
		& \multicolumn{2}{l|}{Min Logit}
		& $0.625\pm0.006$ & $0.749\pm0.005$ & $0.763\pm0.005$
		& $0.604\pm0.028$ & $0.588\pm0.045$ & $0.588\pm0.046$\\
		\midrule

		\multicolumn{3}{l|}{P(True)}
		& $0.730\pm0.005$ & $0.745\pm0.005$ & $0.720\pm0.006$
		& $0.406\pm0.026$ & $0.510\pm0.052$ & $0.361\pm0.017$\\
		\midrule

		\multirow{4}{*}{\rotlabel{Linear Probe\\@ Hook Point\\/ Token}}
		& \multirow{2}{*}{End of Prompt}
		& mlp\_out
		& $0.786\pm0.004$ & $0.729\pm0.005$ & $0.783\pm0.005$
		& $0.722\pm0.035$ & $0.663\pm0.013$ & $0.729\pm0.029$\\
		&
		& resid\_post
		& $0.762\pm0.005$ & $0.720\pm0.005$ & $0.749\pm0.005$
		& $0.711\pm0.038$ & $0.682\pm0.016$ & $0.736\pm0.021$\\
		\cmidrule(l){2-3}
		& \multirow{2}{*}{\shortstack{Generation of\\Final Token}}
		& mlp\_out
		& $0.693\pm0.006$ & $0.678\pm0.005$ & $0.750\pm0.006$
		& $0.661\pm0.024$ & $0.580\pm0.029$ & $0.617\pm0.046$\\
		&
		& resid\_post
		& $0.679\pm0.006$ & $0.692\pm0.005$ & $0.730\pm0.006$
		& $0.673\pm0.015$ & $0.569\pm0.041$ & $0.621\pm0.047$\\
		\midrule

		\multirow{4}{*}{\rotlabel{MLP Probe\\@ Hook Point\\/ Token}}
		& \multirow{2}{*}{End of Prompt}
		& mlp\_out
		& $0.815\pm0.004$ & $0.806\pm0.004$ & $0.805\pm0.004$
		& $0.730\pm0.029$ & $0.669\pm0.040$ & $0.734\pm0.032$\\
		&
		& resid\_post
		& \underline{$0.826\pm0.004$}
		& $\mathbf{0.839\pm0.004}$
		& \underline{$0.820\pm0.004$}
		& \underline{$0.746\pm0.039$}
		& $\mathbf{0.709\pm0.025}$
		& \underline{$0.752\pm0.027$}\\
		\cmidrule(l){2-3}
		& \multirow{2}{*}{\shortstack{Generation of\\Final Token}}
		& mlp\_out
		& $0.754\pm0.005$ & $0.775\pm0.004$ & $0.777\pm0.005$
		& $0.658\pm0.021$ & $0.616\pm0.036$ & $0.592\pm0.017$\\
		&
		& resid\_post
		& $0.767\pm0.005$ & $0.815\pm0.004$ & $0.803\pm0.005$
		& $0.687\pm0.024$ & $0.659\pm0.020$ & $0.613\pm0.037$\\
		\midrule

		\multicolumn{2}{l}{\multirow{2}{*}{NeuroActiSep}}
		& Linear
		& $0.802\pm0.004$ & $0.742\pm0.005$ & $0.768\pm0.005$
		& $0.737\pm0.033$ & $0.671\pm0.024$ & $0.744\pm0.014$\\
		&
		& MLP
		& $\mathbf{0.843\pm0.004}$
		& \underline{$0.833\pm0.004$}
		& $\mathbf{0.830\pm0.004}$
		& $\mathbf{0.769\pm0.026}$
		& \underline{$0.701\pm0.019$}
		& $\mathbf{0.761\pm0.035}$\\
		\bottomrule
	\end{tabular*}
    \endgroup
\end{table*}

Across all three language models, linear probes trained on the selected neurons outperform probes trained on the final-layer residual stream and the MLP output. The bottom-ranked neurons perform close to chance, and the high-variance, low-scored control remains substantially below that of the selected neurons. Nevertheless, the mean AUROC over uniform random draws is higher than that of the top-ranked neurons
across all three models.

Motivated by these observations and for further investigation, we construct the aggregated set $\mathcal{S}$ and evaluate NeuroActiSep on other factual QA datasets. In the following subsections, we investigate whether the advantage over internal representations transfers beyond the neuron selection dataset, how performance varies with layer depth and neuron budget, and whether random neurons remain competitive in sparse regimes, where aggregating information from thousands of neurons is less likely to dominate.

\subsection{Performance of NeuroActiSep on Evaluation Datasets}\label{results}
For each LLM, we use its constructed set $\mathcal{S}_{d_{model}}$ to train a NeuroActiSep classifier for hallucination detection on the evaluation datasets. In addition to scores collected and computed during generation, including perplexity, mean token probability, mean
logit, and minimum logit, we compare against the following baselines:

\paragraph{P(True)} To assess a model's awareness of its knowledge boundary, we use Kadavath et al.~\cite{kadavath2022languagemodelsmostlyknow}'s idea to ask the LLM whether its generated answer is correct. Each model is prompted twice without including any brainstormed ideas, using both orderings of the ``(A) True / (B) False'' options to control for bias.
\paragraph{Linear Probes} Inspired by Orgad et al.~\cite{orgad_llms_2025}, we train separate linear probes on internal representations at the final layer. We consider the MLP output (\texttt{mlp\_out}) and residual stream (\texttt{resid\_post}) at two token positions: the final prompt token and the state leading to the final generated token.
\paragraph{MLP Probes} Similar to Azaria and Mitchell~\cite{azaria_internal_2023}, we train three-layer MLP neural networks on representations from the last transformer layer using the same two hook points and token positions as in the linear baselines.

Table~\ref{tab:results} presents the results. Reported AUROCs for TriviaQA are over 10,000 stratified bootstrap resamples of the test split described in Section \ref{subsec:data}. TruthfulQA results are size-weighted cross-validation over five folds. Despite using neuron locations chosen from a different dataset, the NeuroActiSep MLP is the highest-ranking method in four of the six model--dataset combinations and the second-highest in the remaining two. This indicates the selected neurons contain signals correlated with factual hallucination on questions other than the dataset on which they were identified. Also, the performance of the linear NeuroActiSep probe matches or exceeds the performance of the probe on the residual stream in five of the six combinations and remains close to it for Qwen on TruthfulQA. These results extend the corresponding observation from the neuron selection dataset.

Fig.~\ref{fig:analysis} shows the layer distribution of set $\mathcal{S}_{d_{model}}$ for all models. A recurring theme among the three models is that most neurons are selected from the upper layers. In Mistral and Llama, the number of selected neurons begins to increase around the middle layers, while in Qwen, neurons are considerably concentrated in the upper layers. 

Nevertheless, the concentration of selected neurons in particular layers does not by itself establish the necessity of these layers for predictive performance. Moreover, using the last layer may result in suboptimal performance for the probes trained on the internal states. Therefore, we use the neuron selection dataset as validation data to find the layer $\ell^{*}$ with the best detection performance over three-fold cross-validation. Then, we repeat the fold-specific neuron selection and aggregation, retaining a budget of $d_{\mathrm{model}}$, while restricting eligible neurons to layers up to and including the selected layer $\ell^{*}$. This naturally puts NeuroActiSep under constraint, as the neurons are selected up to the best layer found using the internal states. Regardless, the results in Table~\ref{tab:optimal_layer} are consistent with our previous observations. Layer-restricted NeuroActiSep probes match or exceed their corresponding probes on internal representations in every TriviaQA comparison and in five of the six TruthfulQA comparisons.

\begin{table}[!t]
\centering
\caption{Selected $\ell^{*}$ for each evaluated language model}
\label{tab:l_star}
\begingroup
	\footnotesize
	\begin{tabular*}{\columnwidth}
    {@{\extracolsep{\fill}}c|ccc}
    \toprule
    Selected Layer &Mistral&Qwen&Llama\\
    \midrule
    $\ell_{res}^{*}$&25&21&31\\
    \midrule
    $\ell_{mlp}^{*}$&22&22&26\\
    \bottomrule
    \end{tabular*}
    \endgroup
\end{table}

\begin{table}[!b]
	\centering
	\caption{AUROC of MLP probes at layers selected using the neuron selection dataset as validation data.}
	\label{tab:optimal_layer}
	\begingroup
	\footnotesize
	\begin{tabular*}{\columnwidth}
	{@{\extracolsep{\fill}}c|ccc}
		\toprule
		Feature & Mistral & Qwen & Llama\\
		\midrule

		\multicolumn{4}{c}{TriviaQA}\\
		\midrule
		\shortstack{Selected neurons\\$\leq\ell_{\mathrm{res}}^\star$}
		& $\mathbf{0.828\pm0.004}$
		& $0.846\pm0.003$
		& $0.821\pm0.004$\\
        \cmidrule{1-1}
		\shortstack{Residual stream\\at $\ell_{\mathrm{res}}^\star$}
		& $0.826\pm0.004$
		& $0.836\pm0.004$
		& $0.818\pm0.004$\\
        \cmidrule{1-1}
		\shortstack{Selected neurons\\$\leq\ell_{\mathrm{mlp}}^\star$}
		& $\mathbf{0.828\pm0.004}$
		& $\mathbf{0.850\pm0.003}$
		& $\mathbf{0.835\pm0.004}$\\
        \cmidrule{1-1}
		\shortstack{MLP output\\at $\ell_{\mathrm{mlp}}^\star$}
		& $0.815\pm0.004$
		& $0.845\pm0.004$
		& $0.811\pm0.004$\\

		\midrule
		\multicolumn{4}{c}{TruthfulQA}\\
		\midrule
		\shortstack{Selected neurons\\$\leq\ell_{\mathrm{res}}^\star$}
		& $0.754\pm0.036$
		& $\mathbf{0.740\pm0.030}$
		& $0.776\pm0.025$\\
        \cmidrule{1-1}
		\shortstack{Residual stream\\at $\ell_{\mathrm{res}}^\star$}
		& $\mathbf{0.765\pm0.021}$
		& $0.727\pm0.026$
		& $0.749\pm0.040$\\
        \cmidrule{1-1}
		\shortstack{Selected neurons\\$\leq\ell_{\mathrm{mlp}}^\star$}
		& $0.762\pm0.011$
		& $0.732\pm0.024$
		& $\mathbf{0.782\pm0.024}$\\
        \cmidrule{1-1}
		\shortstack{MLP output\\at $\ell_{\mathrm{mlp}}^\star$}
		& $0.742\pm0.020$
		& $0.720\pm0.019$
		& $0.744\pm0.031$\\

		\bottomrule
	\end{tabular*}

	\endgroup
\end{table}

\subsection{Analysis Across Layer Depth}\label{subsec:depth}
We conduct a depth-restricted analysis using the neuron selection dataset to examine how performance changes as neurons from progressively deeper layers become available. For each depth cutoff $\ell$, neuron selection is restricted to layers up to and including $\ell$. The top row of Fig.~\ref{fig:sensitivity} reports the three-fold cross-validation results under three fold-specific selection strategies: \emph{global}, which selects the highest-ranked $K$ neurons among all neurons up to layer $\ell$, where $K$ is fixed to 1\% of the model's total feed-forward neurons; \emph{layer-balanced}, which distributes a matched total number of selected neurons evenly across the eligible layers; and \emph{fixed-1\%-per-layer}, which selects the top 1\% of neurons independently from each eligible layer.

Hallucination-predictive information appears to be limited in the lower layers and becomes increasingly accessible as neurons from the middle layers are included. The \emph{fixed-1\%-per-layer} strategy has a feature budget half that of the \emph{global} strategy at the midpoint of the language model, but it performs comparably to the other two strategies. This competitive performance of the \emph{fixed-1\%-per-layer} strategy, despite its smaller feature budget at intermediate depths, and the plateaus in the depth curves suggest that most of the signal is already available from a sparser set of neurons before the final layers are introduced. The concentration of selected neurons in the middle-to-late layers, despite the limited gains after the curves plateau, suggests that many highly ranked neurons in later layers may encode overlapping predictive information.

\subsection{Sensitivity to Neuron Budget}\label{subsec:sensitivity}
\begin{figure*}[t]
	\centering
	\includegraphics[width=0.90\textwidth,
    trim={1mm 1mm 1mm 1mm},
    clip]{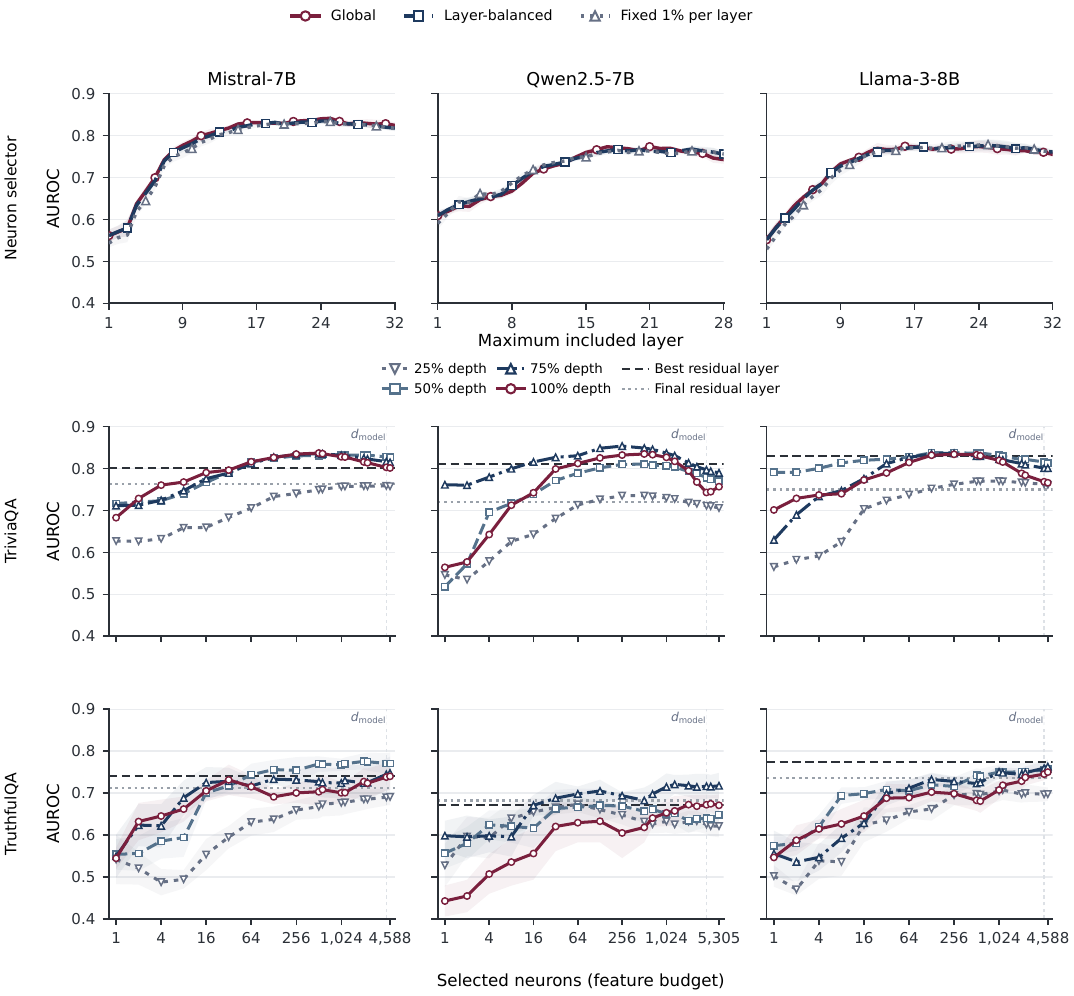}
	\caption{AUROC of linear NeuroActiSep on the neuron selection dataset for each depth cutoff and analysis of sensitivity to neuron budget up to $\mathcal{S}_{1\%}$}
	\label{fig:sensitivity}
\end{figure*}

Results from Section~\ref{subsec:depth} motivate us to investigate the sensitivity of NeuroActiSep to neuron budget and included layers. We examine how performance changes jointly with neuron budget and maximum selection depth. For each model, we consider logarithmic budgets ranging from a single neuron to the $\mathcal{S}_{1\%}$. We repeat the source fold-specific ranking and aggregation separately at four depth cutoffs, restricting eligible neurons to the first 25\%, 50\%, 75\%, or 100\% of the model's layers. Within each cutoff, the evaluated budgets form nested prefixes of the resulting ranking. As in the main transfer experiment, training samples from the target dataset are used to train the linear classifier but not to select or rank neurons. The second and third rows of Fig.~\ref{fig:sensitivity} show the results for TriviaQA and TruthfulQA, respectively. 

The sensitivity results show that at least one selected-neuron configuration outperforms the final-layer residual stream in all six model--dataset combinations. It also exceeds the linear probe on the residual stream at $\ell_{res}^{*}$ in five combinations, with Llama on TruthfulQA as the only exception. On TriviaQA, performance generally increases rapidly over the first few hundred neurons and then plateaus as additional neurons are introduced. The strongest curves peak at 512 neurons for Mistral and at 256 neurons for both Qwen and Llama, substantially below their respective $d_{model}$. Notably, with only 16 selected neurons, the linear NeuroActiSep probe numerically outperforms the linear classifier trained on the last-layer residual stream on TriviaQA. The TruthfulQA curves are less smooth and reach their maxima at larger, model-dependent budgets, but with 256 neurons they perform comparably to the last-layer residual stream. These results show that the largest neuron set is not generally the most effective one: a comparatively small prefix of the source ranking can retain, and sometimes improve, predictive performance.

The results for maximum included depth are also notable. Across both datasets, the full-depth ranking is optimal only for Mistral on TriviaQA; in the other five model--dataset combinations, a ranking restricted to the first 50\% or 75\% of layers performs best. Taken together with Table~\ref{tab:l_star}, these results suggest that intermediate-to-late layers already make hallucination-predictive information available to probes. Overall, our observations suggest that admitting additional layers or neurons does not produce monotonic improvements.

\begin{figure*}[!ht]
	\centering
	\includegraphics[width=0.90\textwidth,
    trim={1mm 1mm 1mm 1mm},
    clip]{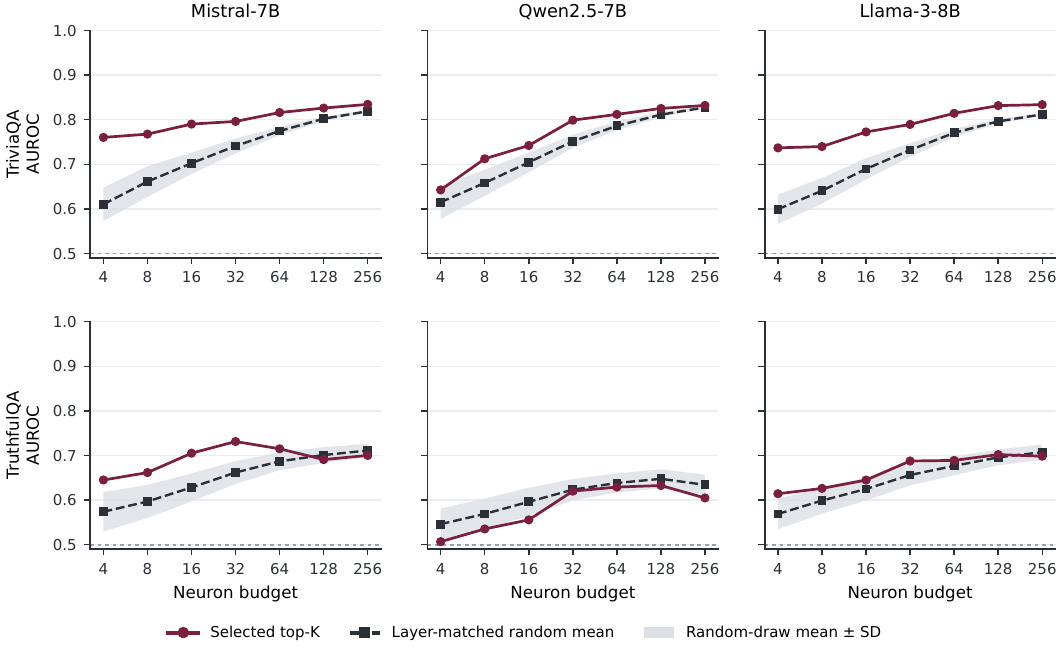}
	\caption{Comparison of selected neurons with random draws of neurons from the same layers as the selected set using linear NeuroActiSep}
	\label{fig:sparse_random}
\end{figure*}

\subsection{Effect of Neuron Selection}\label{subsec:selection_effect}

We have provided empirical evidence that probes trained on sparse sets of feed-forward neurons retain predictive performance and can outperform probes trained on the residual stream. However, such an observation does not on its own justify the effectiveness of selection. As we observed in Table~\ref{tab:counterfact}, randomly drawn neuron sets perform competitively with the selected set. Therefore, we analyze whether this competitiveness also holds in sparser regimes.

The results in Fig.~\ref{fig:sparse_random} compare prefixes of the aggregated ranking of neurons from the entire model with layer-matched random controls. For each budget
$K\in\{4,8,16,32,64,128,256\}$, we generate 250 random neuron sets containing exactly the same number of neurons from each layer as the corresponding selected top-$K$ set. Neurons are sampled uniformly without replacement within each layer. We then train the same linear NeuroActiSep classifier using the signed projected features. In Fig.~\ref{fig:sparse_random}, the random-control curve reports the mean AUROC across the 250 draws, while its shaded region represents the sample standard deviation across random-set identities.

On TriviaQA, the selected neurons outperform the layer-matched random mean for every model and budget. The advantage is largest in the sparsest settings. The results on TruthfulQA are more dependent on the model and feature budget. For Qwen, the selected sets do not outperform the layer-matched random mean at any evaluated budget, and their performance stays mostly within one standard deviation of the random-draw mean. However, in the other two evaluated models, the selected neurons outperform the random mean for Mistral through $K=64$ and for Llama with slight improvement through $K=128$.

These results show that NeuroActiSep can identify neuron identities that are more informative than random neurons from the same layers in specific datasets under sparse budgets. Comparison with the layer-matched random control shows NeuroActiSep provides its clearest advantage at very small feature budgets, particularly on TriviaQA. The diminishing advantage as the budget grows is consistent with predictive information being distributed across many neurons, allowing sufficiently large random sets to recover part of the same signal. This observation is in line with recently reported results by Vankadaru et al. \cite{vankadaru2026readablecontrollableneuronlevelevidence}, which show that selection helps in extremely sparse budgets, while random sets containing a few hundred neurons can recover much of the signal correlated with hallucination. Thus, the selected neurons should be interpreted as efficient
coordinates for constructing a compact predictive representation, rather
than as the only neurons carrying information related to hallucination.
 
\section{Limitations and Conclusion}\label{sec:conclusion}
In this study, we proposed the NeuroActiSep score for ranking neurons and the NeuroActiSep classifier for detecting hallucinations at the final prompt token. We utilized this method for an empirical study on using feed-forward neurons for hallucination detection during the first forward pass of the language model. Our results demonstrated the feasibility of identifying very small sets of neurons correlated with factual hallucinations and using them to train classifiers on out-of-distribution factual questions. These findings do not establish that the selected neurons causally produce or prevent hallucinations, as we did not conduct causal intervention. Therefore, further mechanistic studies are needed to investigate the causal relationship between these neurons and hallucination in language models.

Moreover, these findings should be interpreted within several limitations. The proposed method depends on the quality and content of the neuron selection dataset. Since this study focuses on factual question answering, we did not investigate whether the selected neurons transfer to other domains such as mathematics or programming; applying the method elsewhere may require constructing a selection dataset tailored to each target domain. While probes on sparse neuron sets have fewer input features than probes on the residual stream, we did not analyze the efficiency gains from using sparse sets of feed-forward neurons. Whether such sparse probes offer practical efficiency benefits in latency, memory, or storage relative to a probe on the residual stream remains to be determined. In addition, our experiments cover models with 7 to 8 billion parameters. Future work could examine substantially larger or smaller models and compare how sparse the hallucination signal is at the neuron level across scales. 

\section*{Acknowledgment}
We would like to acknowledge the use of LLM-based tools for assistance in the preparation of this paper. Grammarly Writing Assistant was used to correct spelling, grammar, and punctuation errors. GPT-5.6 Sol was used for coding the plots in Fig.~\ref{fig:sensitivity} and Fig.~\ref{fig:sparse_random}. The authors maintain full accountability and responsibility for the content of this paper.

\bibliographystyle{IEEEtran}
\bibliography{references}

\end{document}